\pdfoutput=1
\documentclass[11pt]{article}

\usepackage[]{EMNLP2022}

\usepackage{times}
\usepackage{latexsym}

\usepackage[T1]{fontenc}
\usepackage[utf8]{inputenc}

\usepackage{microtype}

\usepackage{inconsolata}

\usepackage{algorithm}
\usepackage{algorithmic}

\usepackage[textsize=scriptsize]{todonotes}

\usepackage{cleveref}

\crefname{section}{§}{§§}

\usepackage{multicol}
\usepackage{multirow}
\usepackage{newfloat}
\usepackage{listings}
\floatstyle{ruled}
\newfloat{listing}{tb}{lst}{}
\floatname{listing}{Listing}

\title{Mixed-modality Representation Learning and Pre-training for \\ Joint Table-and-Text Retrieval in OpenQA}

\author{Junjie Huang$^{1\dag}$\thanks{\ \ \ Indicates equal contribution}, Wanjun Zhong $^{2*}$\thanks{\ \ \ Work is done during internship at Microsoft Research Asia.}, Qian Liu$^{1\dag}$,
\textbf{Ming Gong$^3$, Daxin Jiang$^3$, Nan Duan $^4$}\\
	$^1$ Beihang University \quad $^2$ Sun Yat-sen University \\
    $^3$ Microsoft STC Asia \quad $^4$ Microsoft Research Asia\\
    {\tt \{huangjunjie, qian.liu\}@buaa.edu.cn} \\
    {\tt \{mingo, djiang, nanduan\}@microsoft.com} \\
	{\tt zhongwj25@mail2.sysu.edu.cn} 
}

\begin{document}
\maketitle
\begin{abstract}

Retrieving evidences from tabular and textual resources is essential for open-domain question answering (OpenQA), which provides more comprehensive information. 
However, training an effective dense table-text retriever is difficult due to the challenges of table-text discrepancy and data sparsity problem. 
To address the above challenges, we introduce an optimized \textbf{O}penQA \textbf{T}able-\textbf{\textsc{Te}}xt \textbf{R}etriever (\textsc{OTTeR}) to jointly retrieve tabular and textual evidences. 
Firstly, we propose to enhance mixed-modality representation learning via two mechanisms: modality-enhanced representation and mixed-modality negative sampling strategy.
Secondly, to alleviate data sparsity problem and enhance the general retrieval ability, we conduct retrieval-centric mixed-modality synthetic pre-training. 
Experimental results demonstrate that \textsc{OTTeR} substantially improves the performance of table-and-text retrieval on the OTT-QA dataset. 
Comprehensive analyses examine the effectiveness of all the proposed mechanisms.
Besides, equipped with \textsc{OTTeR}, our OpenQA system achieves the state-of-the-art result on the downstream QA task, with 10.1\% absolute performance gain in terms of the exact match over the previous best system.
\footnote{All the code and data are available at \url{https://github.com/Jun-jie-Huang/OTTeR}.}

\end{abstract}

\section{Introduction}\label{sec:intro}
\begin{figure}[t]
     \centering
     \includegraphics[width=0.475\textwidth]{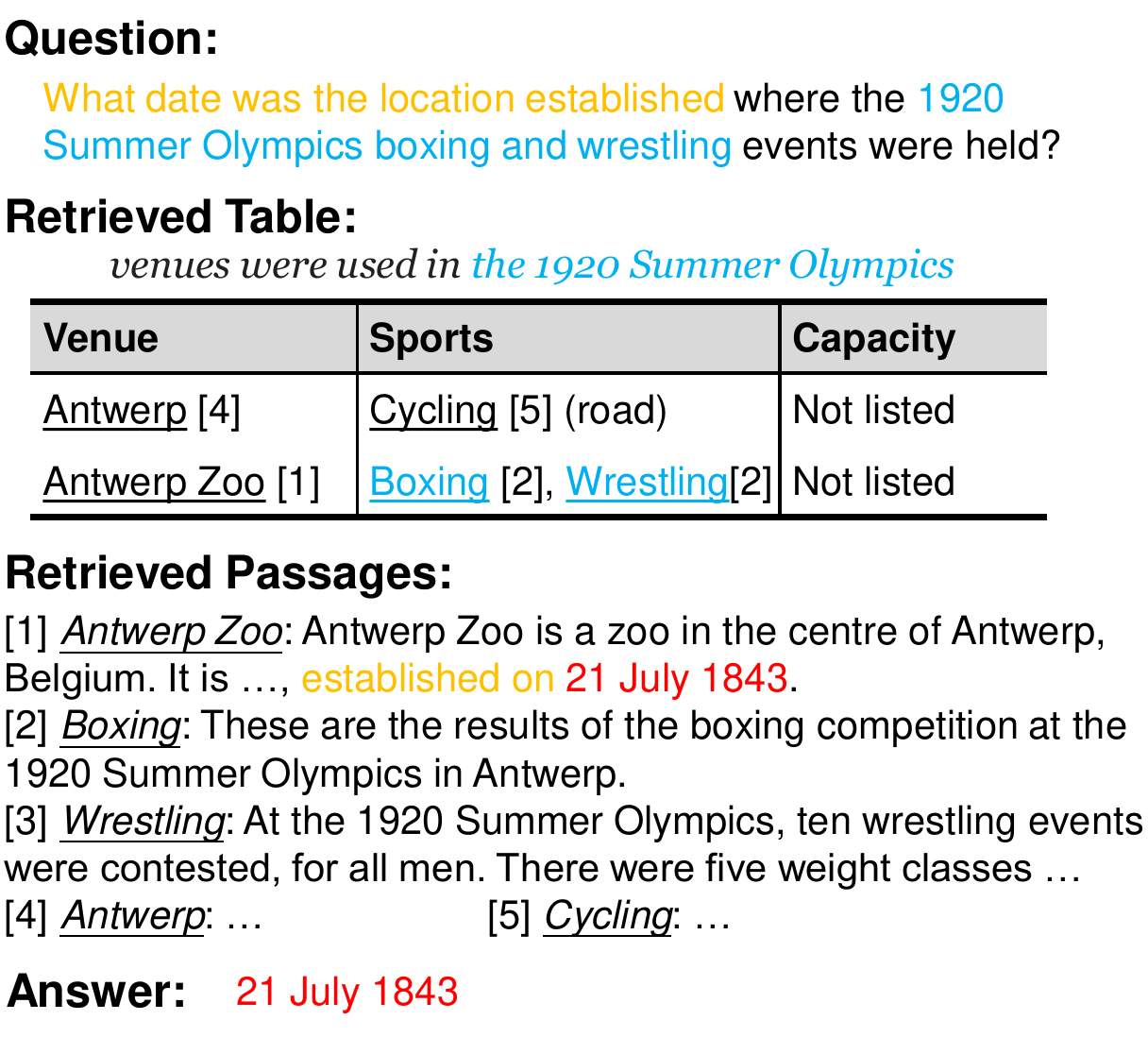}
     \caption{An example of the open question answering over tables and text. 
     Highlighted phrases in the same color indicate evidence pieces related to the question in each single modality. The answer is marked in red.}
    %  Words in the same color denote evidences corresponding to questions, which locate separately in heterogeneous resources.}
     \label{fig:example}
    %  \vspace{-5mm}
\end{figure}

Open-domain question answering \cite{Joshi2017TriviaQA,  dunn2017searchqa, Lee2019LatentRF} aims to answer questions with evidence retrieved from a large-scale corpus.
% aims to answer questions by retrieving supported evidence from a large-scale corpus and making inference over it.
% from a large pool of Wikipeida knowledge.  
The prevailing solution follows a two-stage framework \cite{Chen2017ReadingWT}, where a \textit{retriever}
first retrieves relevant evidences
% first selects a list of \textit{k} passages as evidences 
and then a \textit{reader} extracts answers from the evidences. 
Existing OpenQA systems \cite{Lee2019LatentRF, karpukhin2020dense, Mao2021GAR} have demonstrated great success in retrieving and reading passages. 
% Most approaches focus on questions whose answers reside in single modal evidences, such as free-form text \cite{Xiong2021mdr} or semi-structured tables \cite{Herzig2021nqtables}.
However, most approaches are limited to questions whose answers reside in single modal evidences, such as free-form text \cite{Xiong2021mdr} or semi-structured tables \cite{Herzig2021nqtables}.
% In fact, there exist complex questions that require aggregating evidences from multi-modal resources. 
% In fact, a significant amount of questions can not be answered with single-modal evidences. 
% As the example in Figure \ref{fig:example}, to answer the question, a model first locates the row of ``Antwerp Zoo" in the table, which is the boxing and wrestling venue of 1920 Olympics. Then it reads the relevant passage of Antwerp Zoo to find the answer ``21 July 1843". 
% a model requires to find the evidence that ``\textit{Antwerp Zoo held boxing and wrestling events in 1920 Olympics}" from the table and the evidence that ``\textit{Antwerp Zoo is established on \textbf{21 July 1843}}" from the passage, and then perform multihop reasoning over the two evidences.
% However, in a real world scenario, a significant amount of knowledge resides both in free-form passages and semi-structured tables. % However, in a real world scenario, a great many questions require aggregating evidences from multi-modal 
% Note that the table and the passage contain non-overlapping information. 
% Therefore, answering open-domain questions over both tables and texts is essential and challenging since systems should be capable to handle the heterogeneous data. 
% Answering such questions require access of both tables and passages
% In fact, there exist complex questions that requires 
However, solving many real-world questions requires aggregating heterogeneous knowledge (e.g., tables and passages), because massive amounts of human knowledge are stored in different modalities.
% In a real-world scenario, a huge amount of human knowledge is distributed over heterogeneous resources, e.g., tables and passages.
% Besides, a large number of questions are required to be answered by aggregating heterogeneous knowledge. 
% Using homogeneous information alone might lead to severe coverage problems. 
As the example shown in Figure \ref{fig:example}, the supporting evidence for the given question resides in both the table and related passages.
% the question can be answered in the absence of either the table or the passages. 
Therefore, retrieving relevant evidence from heterogeneous knowledge resources involving tables and passages is essential for advanced OpenQA, which is also our focus.
% is an essential preliminary to OpenQA. 
% Answering such questions require a 
% In this paper, we focus on jointly retrieving cross-modal evidences of tables and text for OpenQA. 

% We study the problem of OpenQA over tables and text in this paper (Figure \ref{fig:pipeline}). Specially we focus on the retriever to jointly extract the cross-modal evidences. 
% % 怎么引到说我们不用iterative的方法,我们先把passage link到passages上面, 需要解释我们没有用iterative retriever吗？
% Most recent works \cite{Li2021HopRetriever, Xiong2021mdr} retrieve multiple evidences in an iterative manner. However, this might not be the best choice due to the discrepancy of two modalities. 
% % As tables often contain large quantities of entities, it's a natural way to link passages to the entities in tables as a complementary source and retrieve a unit of table and text at a time. 
% A straightforward way is to align passages to the entities in tables and view the fused block as a basic unit of retrieval. In this case, evidences from two resources complement each other and iterative retriever is no more necessary.
% % The major challenges of this task lie in two directions.

\begin{figure*}[thbp]
     \centering
     \includegraphics[width=0.97\textwidth]{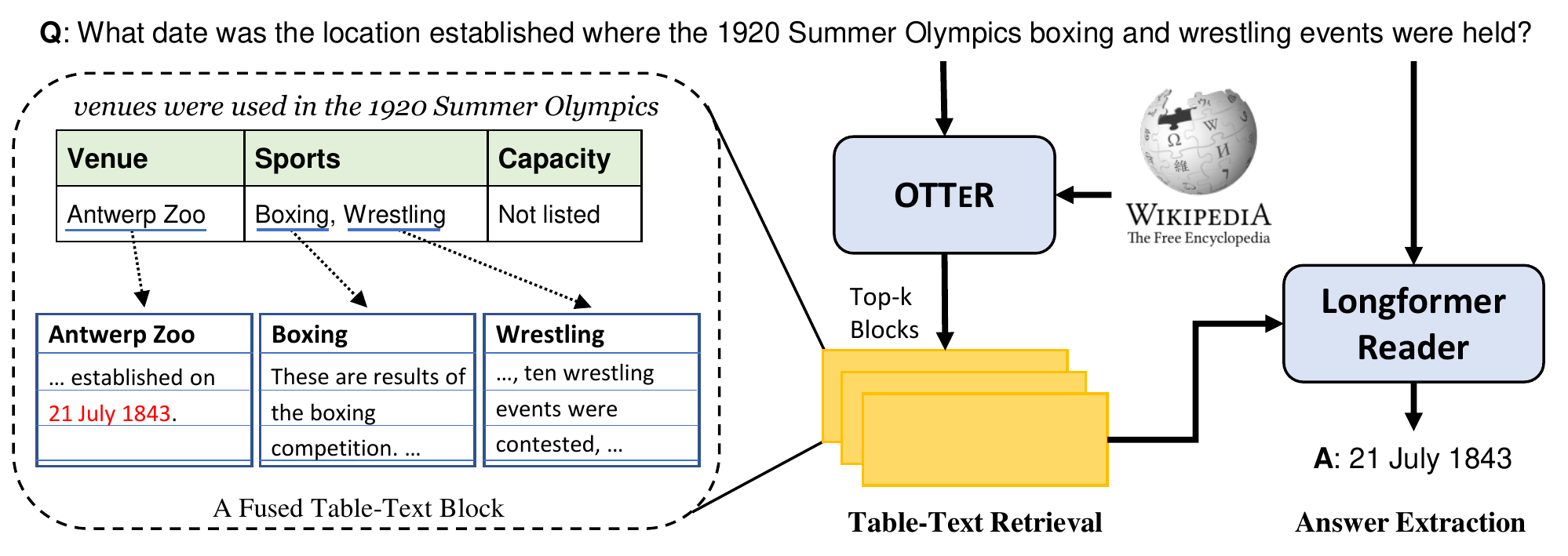}
     \caption{The framework of the overall OpenQA system. 
    %  It first constructs table-text pairs by linking passages to the cells in the table. 
     It first jointly retrieves top-k table-text blocks with our \textsc{OTTeR}. 
     Then it answers the questions from the retrieved evidence with a reader model. }
     \label{fig:pipeline}
    %  \vspace{-3mm}
\end{figure*}

There are two major challenges in joint table-and-text retrieval: 
% lie in two directions. 
(1) There exists the discrepancy between table and text, which leads to the difficulty of jointly retrieving heterogeneous knowledge and considering their cross-modality connections; 
% Therefore, it is hard to retrieve two modalities of information simultaneously and consider their connections.
% The representations of fused blocks should involve information of the two parts. However, it is difficult to represent two modalities in a single vector with less information loss.
% How to effectively represent the two modalities of data. 
% One of the major challenges to train a joint table and text retriever is. First,
% Second, it is expensive to acquire large-scale training pairs of question and fused blocks for OpenQA retrievers. OTT-QA \cite{chen2020ottqa} is the largest dataset for OpenQA over text and table with $45K$ annotated questions. However, it is still insufficient to cover all topics of questions and tables.
(2) The data sparsity problem is extremely severe because
training a joint table-text retriever requires large-scale supervised data to cover all targeted areas, which is labourious and impractical to obtain. 

% Second, how to acquire more training pairs of question and fused segment for OpenQA. 

In light of this two challenges, we introduce an optimized \textbf{O}penQA \textbf{T}able-\textbf{\textsc{Te}}xt \textbf{R}etriever, dubbed \textsc{OTTeR}, which utilizes mixed-modality dense representations to jointly retrieve tables and text.
% To address the above two challenges, we propose three novel solutions.
Firstly, to model the interaction between tables and text, we propose to enhance mixed-modality representation learning via two novel mechanisms: modality-enhanced representations (MER) and mixed-modality hard negative sampling (MMHN).
% For the problem of table-text discrepancy, we apply the modality-enhanced representations (MER) and mixed-modality hard negative sampling (MMHN) to enhance cross-modal interactions. 
MER incorporates fine-grained representations of each modality to enrich the semantics.
MMHN utilizes table structures and creates hard negatives by substituting fine-grained key information in two modalities, to encourage better discrimination of relevant evidences. 
Secondly, to alleviate the data sparsity problem and empower the model with general retrieval ability, we propose a retrieval-centric pre-training task with a large-scale synthesized corpus, which is constructed by automatically synthesizing mixed-modal evidences and reversely generating questions by a BART-based generator.

Our primary contributions are three-fold:
% \begin{itemize}
    % \item 
    (1) We propose three novel mechanisms to improve table-and-text retrieval for OpenQA, namely modality-enhanced representation, mixed-modality hard negative sampling strategy, and mixed-modality synthetic pre-training.
    % \item 
    (2) Evaluated on OTT-QA, \textsc{OTTeR} substantially improves retrieval performance compared with baselines. Extensive experiments and analyses further examine the effectiveness of the above three mechanisms. 
    % Experimental results show that the three strategies are effective to improve the retrieval performance.
    % \item 
    (3) Equipped with \textsc{OTTeR}, our OpenQA system significantly surpasses previous state-of-the-art models with 10.1\% absolute improvement in terms of exact match.
% \end{itemize}

% We also propose to synthesize the pre-training corpus by automatically synthe-size pseudo evidence chains from the Wikipedia corpus con-sisting of passages and tables, and generating correspondingmulti-hop questions with a BART-based generation model.The pre-training process shows its effectiveness by improv-ing the performance of evidence chain extraction.

% \section{Methodology}
% Here, we present \textsc{OTTeR}, a joint table-text retriever for OpenQA. We first introduce the background of fused table-text blocks and dual-encoder architecture for dense retrieval. Then we describe the three novel strategies in \textsc{OTTeR}. 

% \vspace{-2mm}
\section{Background} \label{sec:preliminary}
% \subsubsection{Task Definition}
% In this paper, we deal with the task of jointly retrieving table-text evidences for given natural language questions. 
% The retrieval corpus consists of a set of table-text blocks $C_B=\{b_1, b_2, ..., b_B\}$, where $b_i$ is pre-fused with a table segment $t_i$ and a passage segment $p_i$.
% Given a question $q$ as input, the task aims to retrieve $k$ blocks $B=\{b_1, b_2, ..., b_k\}$ that can potentially answer question $q$. 
% sorted by their relevance to question $q$
\subsection{Problem Formulation}
% The task of open-domain question answering over tables and text is defined as follows.
The task of OpenQA over tables and text is defined as follows.
Given two corpus of tables $C_T=\{t_1, ..., t_T\}$ and passages $C_P=\{p_1, ..., p_P\}$, the task aims to answer question $q$ by extracting answer $a$ from the knowledge resources $C_P$ and $C_T$. 
The standard system of solving this task involves two components: a \textit{retriever} that first retrieves relevant evidences $c \subset C_T \cup C_P$, and a \textit{reader} to extract $a$ from the retrieved evidence set.
% evidences $c \in C_T \cup C_P$,

% where $c$ can be either tables or passages. 

% The task of open-domain question answering over tables and text is defined as follows. Given a question $q$ and a corpus of table-text blocks $C$, a \textit{retriever} aims to return a list of $n$ blocks $B=\{b_1, b_2, ..., b_n\}$ sorted by their relevance to question $q$, where a table-text block $b$ consists of a table segment $t$ and a passage segment $p$. Then a \textit{reader} reads the retrieved blocks $B$ and extracts a span that answers question $q$. In this paper, we focus on proposing a table-text retriever to retrieve fused blocks that contain the answer. 

\subsection{Table-and-text Retrieval}
% Since relevant table and passages are not necessarily composed,
In this paper, we focus on table-and-text retrieval for OpenQA. 
% Different from single table-based retrieval and text-based retrieval, table-and-text retrieval considers a table segment and relevant passage segment as a self-contained block, which serves as a basic unit for retrieval. 
% To better align the mixed-modality information in table-and-text retrieval, we follow \citet{chen2020ottqa}, and tackle a table segment and relevant passages as a self-contained block, which serves as a basic unit for retrieval. 
To better align the mixed-modality information in table-and-text retrieval, we follow \citet{chen2020ottqa} and take a table-text block as a basic retrieval unit, which consists of a table segment and relevant passages.
Different from retrieving a single table/passage, retrieving table-text blocks could bring more clues for retrievers to utilize since single modal data often contain incomplete context. 
% The advantage of this method is that it could bring more clues for retrievers to utilize since single modal data often contain incomplete context. 
Figure \ref{fig:pipeline} illustrates table-and-text retrieval and our overall system.
% where a table and relevant passages are considered as a whole.

\subsubsection{Table-Text Block} 
% We first define a table-text block $b$ 
% Since there are no annotated links between the table and passages, we need to pre-align the entities in the table and passages by an entity-linking model.
Since relevant tables and passages do not necessarily naturally coexist, we need to construct table-text blocks before retrieval. 
One observation is that tables often hold large quantities of entities and events. 
Based on this observation and prior work \cite{chen2020hybridqa}, we apply entity linking to group the heterogeneous data. 
% Instead of separately retrieve tables and passages, we apply ``earch fusion" strategy to group relevant heterogeneous data before retrieval as \citet{chen2020ottqa}. 
% This procedure fuses relevant tabular and textual data into a self-contained block, which later serves as a basic unit for retrieval. 
Here we apply BLINK \cite{wu2019blink} to fuse tables and text, which is an effective entity linker and capable to link against all Wikipedia entities and their corresponding passages. 
Given a flat table segment, BLINK returns $l$ relevant passages linked to the entities in table. 
% , which contains a table segment and $m$ relevant passages $\{p_1, ..., p_m\}$ linked to entities in the table by an entity linker. 
% However, as BERT-based encoders \cite{Devlin2019BERT} are incapable to long sequence input, we split a full table into several blocks. 
However, as table size and passage quantity grow, the input may become too long for BERT-based encoders \cite{Devlin2019BERT}. 
Thus, we split a table into several segments to limit the input token number that each segment contains only a single row. 
This setup can be seen as a trade-off to resolve input limit but our approaches are scalable to full tables when input capacity permits.
% with the schema. 
% In this paper, each fused block only contains a single row of a  table and its related passages. 
% In this paper, each block only contains a single row of table, with the linked passages. 
More details about table-text blocks can be found in Appendix \ref{app:block-representatoin}.
% More details about block constructions and representations can be found in Appendix \ref{app:block-representatoin}. \TODO

\iffalse
\subsection{Evaluation Protocol} \label{sec:evaluation-protocal}
A well-recognized metric for information retrieval is the recall at top k ranks (R@k), which is the proportion of relevant items found in the top-k turned instances. 
In this paper, we use two metrics to evaluate retrieval systems. 
One metric is the table recall, where the gold table is the ground-truth table annotated by human. 
However, in table-and-text retrieval, table recall is imperfect as an coarse-grained metric since our basic retrieval unit is a table-text block corresponding to a specific row in the table. 
Therefore we propose a more challenging metric
called block recall at top k ranks, where a fused
block is considered a correct match when it meets two requirements. 
First, it comes from the ground truth table. Second,
it contains the correct answer.
\fi

\section{Methodology}
% In this section, we present \textsc{OTTeR}, a joint table-text retriever for OpenQA. 
We present \textbf{\textsc{OTTeR}}, an \textbf{O}penQA \textbf{T}able-\textbf{\textsc{Te}}xt \textbf{R}etriever.
We first introduce the basic dual-encoder architecture for dense retrieval (\cref{sec:biencoder}). 
We then describe three mechanisms to mitigate the table-text discrepancy and data sparsity problems, i.e., modality-enhanced representation (\cref{sec:MER}), mixed-modality hard negative sampling (\cref{sec:MMHN}), and mixed-modality synthetic pre-training (\cref{sec:PT}).

\subsection{The Dual-Encoder Architecture} \label{sec:biencoder}
The prevailing choice for dense retrieval is the dual-encoder method. 
In this framework, a question $q$ and a table-text block $b$ are separately encoded into two $d$-dimensional vectors by a neural encoder $E(\cdot)$. 
Then, the relevance between $q$ and $b$ is measured by dot product over these two vectors:

\begin{equation}
% \small
    % s(q, b) =  E(q)^{\top} \cdot E(b).
    s(q, b) = \mathbf{q}^{\top} \cdot \mathbf{b} =  E(q)^{\top} \cdot E(b).
\end{equation}
The benefit of this method is that all the table-text blocks can be pre-encoded into vectors to support indexed searching during inference time. 
% This formulation allows all the fused block representations to be pre-encoded
In this work, we initialize the encoder with a pre-trained RoBERTa \cite{Liu2019roberta}, and take the representation of the first \texttt{[CLS]} token as the encoded vector.
When an incoming question is encoded, the approximate nearest neighbor search can be leveraged for efficient retrieval \cite{Johnson2021Faiss}.

\paragraph{Training} 
% The training objective aims to learn a representation space that maps a question and relevant blocks close, while keeping irrelevant ones apart. 
The training objective aims to learn representations by maximizing the  relevance of the gold table-text block and the question.
We follow \citet{karpukhin2020dense} to learn the representations. 
Formally, given a training set of $N$ instances, 
% $\{ (q_i, b_i^{+}, b_{i,1}^{-},...,b_{i,m}^{-})\}^N_i$ of $N$ instances, 
the $i^{th}$ instance $(q_i, b_i^{+},b_{i,1}^{-},...,b_{i,m}^{-})$ consists of a positive block $b_i^{+}$ and $m$ negative blocks $\{b_{i,j}^{-}\}^m_{j=1}$, we minimize the cross-entropy loss as follows:

\begin{small}
% \begin{equation}
\[
    L(q_i, b_i^{+}, \{b_{i,j}^{-}\}^m_{j=1}) = - \log \frac{e^{s(q_i, b_i^+)}}{ e^{s(q_i, b_i^+)} + \sum^{m}_{j=1} e^{s(q_i, b_{i,j}^-)}}.
% \end{equation}
\]
\end{small}
% Typically, one hard negative is leveraged and the rest are in-batch negatives \cite{karpukhin2020dense}.
Negatives are a \textit{hard negative} and $m-1$ \textit{in-batch negatives} from other instances in a mini-batch.
% , which is sampled from a standard retrieval system such as BM25 or a neural network.
% \cite{tanghongyin}
% The rest are from positive blocks which are not paired with $q_i$ in the batch, referred to as \textit{in-batch} negatives.
% Typically, the negatives can be sampled from a noisy distribution, such as candidates from human heuristics \cite{}, BM25 \cite{Nogueira2019PassageBM25} and dense encoders \cite{Xiong2021ance}, or from other examples within the batch \cite{Luan2021SparseDA, Qu2021rocketqa, Yang2021xMoCoCM}.

% \subsection{\textsc{OTTeR}} \label{sec:\textsc{OTTeR}}
% We have discussed two major challenges in training a joint table-text retriever, namely modality discrepancy and data sparsity. 

\subsection{Modality-enhanced Representation}\label{sec:MER}
% Most dense retrievers use a coarse-grained single-modal embedding to represent text, such as the embedding of the first token (\texttt{[CLS]} symbol) \cite{karpukhin2020dense} and the average embedding of all tokens \cite{Zhan2020RepBERT}, which is effective in simple passage retrieval with one modality.
% However, in joint table-and-text retrieval, using only single embeddings may lead to information loss since relevant information resides in both modalities. 
% \cite{Zhan2020RepBERT, Huang2021WhiteningBERT}.
Most dense retrievers use a coarse-grained single-modal representation from either the representation of the \texttt{[CLS]} token or the averaged representations of tokens \cite{Zhan2020RepBERT, Huang2021WhiteningBERT}, which is insufficient to represent cross-modal information. 
To remedy this, we propose to learn modality-enhanced representation (MER) of table-text blocks. 

\begin{figure}[t]
     \centering
     \includegraphics[width=0.475\textwidth]{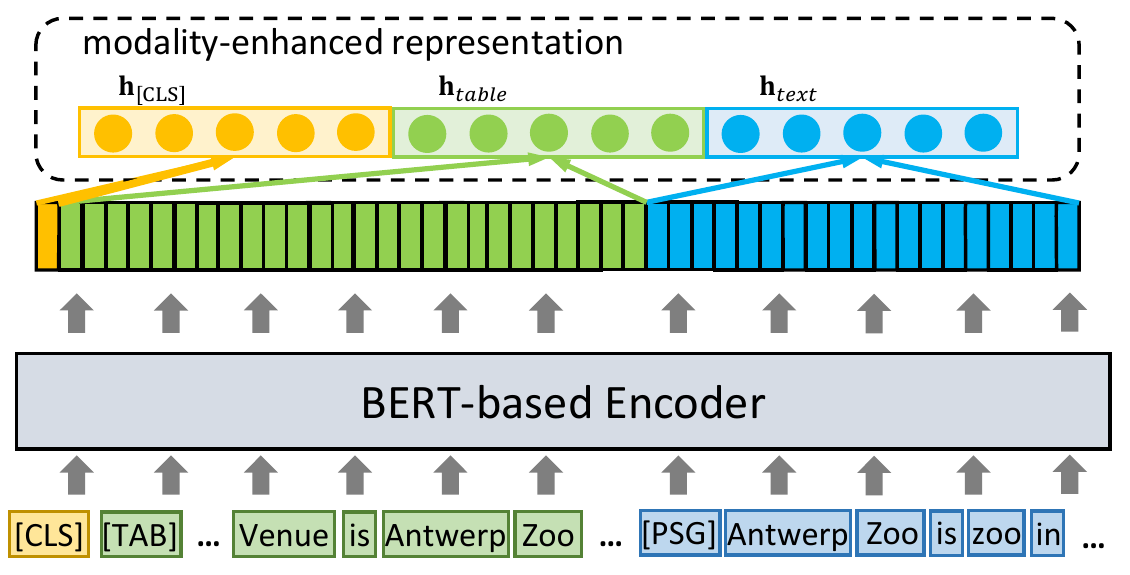}
     \caption{The illustration of modality-enhanced representation in \textsc{OTTeR}. Segments in green and blue denote information of tables and passages respectively. }
     \label{fig:mer-illustration}
    %  \vspace{-3mm}
\end{figure}

As illustrated in Figure \ref{fig:mer-illustration}, instead of using only the coarse representation $\mathbf{h}_{\texttt{[CLS]}}$ at the \texttt{[CLS]} token,
% $\mathbf{b}=\mathbf{h}_{\texttt{[CLS]}}$,  
% MER incorporates tabular and textual embeddings and concatenate the three to represent a fused block.
% The two embeddings are supposed to enhance the semantics of the table and the text in representations, respectively. 
MER incorporates tabular and textual representations ($\mathbf{h}_{table}$ and $\mathbf{h}_{text}$) to enhance the semantics of table and text. 
% The enhanced representation is the concatenation of \texttt{[CLS]}, tabular and textual representations,
Thus, the modality-enhanced representation is 
$\mathbf{b}=[\mathbf{h}_{\texttt{[CLS]}}; \mathbf{h}_{table}; \mathbf{h}_{text} ] $, where $;$ denotes concatenation.

% For a tabular/textual modality, we calculate a unified representation for   in the following ways:  
Given the tokens in a tabular/textual modality, we calculate a representation in the following ways:
% Given all the tokens in one modality, we calculate a unified tabular/textual representation  in the following ways:
% We consider four types of tabular/textual representations: 
(1) \textbf{FIRST}: representations of the beginning token (i.e., \texttt{[TAB]} and \texttt{[PSG]}); 
(2) \textbf{AVG}: averaged token representations; (3) \textbf{MAX}: max pooling over token representations ; (4) \textbf{SelfAtt}: weighted average over token representations where weights are computed by a self attention layer. 
We discuss the impact of different types of MERs in \cref{sec:result-ablation}. 
Our best model adopts FIRST as the final setting.
% embedding to represent uni-modal data. 
To ensure the same vector dimensionality with the enriched representation, we represent the question by replicating the encoded question representation. 
% embedding to ensure the same vector dimensionality. 
% i.e., $\mathbf{q}=[\mathbf{h}_{\texttt{[CLS]}}, \mathbf{h}_{text}, \mathbf{h}_{text} ]$.

\subsection{Mixed-modality Hard Negative Sampling}\label{sec:MMHN}
Prior studies \cite{Nogueira2019PassageBM25, Gillick2019LearningDR} 
% has found that negative sampling strategy is essential in dense retrieval.
have found that hard negative sampling is essential in training a dense retriever.
% Hard negatives are proved important in learning a high-quality dense retriever 
% Previous methods \cite{Yang2021xMoCoCM,Qu2021rocketqa} 
These methods take each evidence as a whole and retrieve the most similar irrelevant one as the hard negative. 
% These methods for single-modality negative sampling take each evidence block as a whole and retrieves the most similar irrelevant evidence block.
% These coarse-grained methods use a whole relevant 
Instead of finding an entire irrelevant block, we propose a \textit{mixed-modality hard negative sampling} mechanism, which constructs more challenging hard negatives by only substituting partial information in the table or text. 
% However, these coarse-grained methods neglect the structures within tables.
% connections between tables and text 
% Therefore, we propose to construct more challenging hard negatives 
% by only substituting partial information in table or text, considering the . 
% either using BM25 \cite{Yang2021xMoCoCM, Luan2021SparseDA}, or using neural models \cite{Qu2021rocketqa, Xiong2021ance}. 
% A traditional negative sampling strategy
% is to sample from top candidates retrieved by BM25 \cite{Yang2021xMoCoCM, Luan2021SparseDA} or by a neural network \cite{Qu2021rocketqa, Xiong2021ance}.
% However, these methods take the fused blocks as a whole but ignore the connections between tables and text.
% ignore the characteristics of joint retrieval over tables and text. 
% Here we propose a simple way to sample hard negative fused blocks by mixing modalities. 
% The idea of constructing mixed-modality hard negatives is to replace only one modality in the positive block and mix the remaining modality with the one in other blocks. 

% The idea of mixed modality hard negative sampling strategy is to
% Specifically, we create hard negatives by substituting either a table row or a passage in the positive block, rather than simply finding a similar entire block.
% to mix one modality in the positive block with the other modality in another block. Intuitively, it creates harder negatives than simply using a random whole block. 
Formally, suppose a positive block  $b^{j+} = (t^{j}, p^{j})$  is from the $j$-th row in the table, the answer $a$ resides in either table segment $t^j$ or passages $p^j$. 
We decide to replace either the table row or the passage depending on where the answer exists.
% on the answer position.
% We consider two situations depending on the answer position.
If $a$ exists in the table row, we construct a hard negative 
$b^{j-} = (t^{k}, p^{j})$ 
by replacing $t^{j}$ with a random row $t^{k}$ in the same table. 
Similarly, if $a$ resides in the passages, we create hard negative $b^{j-} = (t^{j}, p^{k})$ by replacing passages with $p^{k}$ in other blocks.
% A straight-forward way is to randomly take a whole fused block $(q_i,  (t_{i}^{k}, p_{i}^{k}))$ from the $k$-th row of $T_i$ as a negative. 

% \subsection{Pre-training with synthesized corpus}
\subsection{Mixed-modality Synthetic Pre-training}\label{sec:PT}

To alleviate the issue of data sparsity, we propose a mixed-modality synthetic pre-training (MMSP) task.
MMSP enhances the retrieval ability by pre-training on a large-scale synthesized corpus, which involves mixed-modality pseudo training data with \textit{(question, table-text block)} pairs. 
% to pre-train the retriever with mixed-modality synthetic corpus. 
% To alleviate the issue of data sparsity, we propose a pre-training strategy for retrieval with mixed-modality synthetic corpus
% MMSP improves the retrieval ability by pre-training on a large-scale synthesized mixed-modality corpus involving both tables and text. 
% The 
% The key challenge is how to mine positive \textit{(question, evidence)} pairs in an unsupervised manner. 
Here, we introduce a novel way to construct the pseudo training corpus in two steps, including table-text block mining and question back generation. 

% One observation is that the Wikipedia hyperlinks often link explanatory passages to entities in tables, which allows us to parse a huge amount of table and passages pairs.
% The hyperlinks in Wikipedia that link the passages to the entities in tables allow us to parse a huge amount of table and passages pairs, which is the foundation of our corpus construction. 
% Then, with each parsed mixed-modality evidence, the key challenge is how to construct corresponding question for pre-training the retriever. 
% Therefore, we apply a BART-based generation model \cite{lewis2019bart} to automatically generate pseudo question from the parsed evidence, which is called \textit{back-generation}. 

(1) Mine relevant table-text pairs. 
One observation is that Wikipedia hyperlinks often link explanatory passages to entities in tables, which provides high-quality relevant table-text pairs. 
Based on this, we believe Wikipedia is an excellent resource for parsing table-text pairs.
% Based on this, we parse a Wikipedia table with corresponding hyperlink passages as the source of fused blocks. 
% Specially, we first parse a Wikipedia table with Wikipedia passages hyperlinked to it as the source to construct more fused blocks. 
Specially, we select a row in a table, and find corresponding passages with the hyperlinks to form a fused table-text block.
We only keep the first section in each Wikipedia page as it always contains the most important information about the linked entity.
% Next, we automatically write a pseudo question for each fused block. 
(2) Write pseudo questions for fused blocks. 
The questions are expected to not only contain the mixed-modality information from the blocks, but also have good fluency and naturalness. 
Therefore, instead of using template-based synthesizing methods, we use a generation-based method to derive more fluent and diverse questions, which is called \textit{back-generation}.  
% Next, we automatically construct corresponding pseudo question from the parsed evidence block, and we expect the question should contain mixed-modality information and has better fluency. 
% Therefore, instead of using template-based construction method, we use a generation-based method to provide more fluent and diverse questions.
Specially, we use BART$_{base}$ \cite{lewis2019bart} as the backbone of our generator, which is fine-tuned with oracle pairs of \textit{(question, table-text block)} in the OTT-QA training set. 
The input to the generator is a sequence of the flat table and linked passages, and the output is a mixed-modality question. 
Finally, we automatically construct a large-scale pre-training corpus.
We present some examples of generated pseudo questions in Appendix \ref{app:synthetic-corpus}. Finally, we obtain a synthesized corpus with 3M pairs of table-text blocks and pseudo questions.
% More details of corpus synthesizing and examples are presented in Appendix \ref{app:synthetic-corpus}. 

During pre-training, we adopt a similar ranking task where the training objective is the same as described in \cref{sec:biencoder}.
As for negative sampling, we use in-batch negatives and one hard negative randomly sampled from the same table.

% The pre-training task is the same as training a basic dual-encoder dense retriever described in \cref{sec:biencoder}. We use in-batch negatives and one hard negative randomly sampled from the same table.
% Finally, we obtain a synthesized corpus consisting of 3M question-evidence pairs.

% \subsubsection{Pre-train Task}
% The pre-training task is the same as training a basic dual-encoder dense retriever described in \cref{sec:biencoder}. We use in-batch negatives and one hard negative randomly sampled from the same table.

% \subsection{Reader}

% \section{Experiments}
% We experiment on open-domain question answering over tables and text and report results on two subtasks, including table-and-text retrieval and end-to-end question answering. We also give detailed analyses from different perspectives to illustrate the effectiveness of proposed three strategies. 

% \subsection{Dataset and Metrics}

% \vspace{-1mm}
\section{Experiment Settings}
% \vspace{-1mm}
% In this section, we describe the data we used for experiments and the basic setup. 
In this section, we describe the experiment settings  on the task of open-domain question answering over tables and text, and report the performance of our system on the table-and-text retrieval, and downstream question answering.
% and report results on two subtasks, including table-and-text retrieval and downstream question answering. 

\subsection{Dataset} 
% We perform experiments to evaluate our method 
Our system is evaluated 
on the \textbf{OTT-QA} dataset \cite{chen2020ottqa}, which is a large-scale open-domain table-text question answering benchmark. 
Answering questions in OTT-QA requires aggregating multi-modal information from both tables and text. 
% The tables and textual passages in OTT-QA are collected from Wikipedia.
% holding the advantage 
OTT-QA dataset contains over 40K questions with human annotated answers and ground truth evidences.
It also provides a corpus of over 400K tables and 6M passages collected from Wikipedia.
% Moreover, the ground truth tables and passages are not given in OTT-QA. Systems require to retrieve relevant tables and text, which 
% Moreover, OTT-QA requires systems to retrieve relevant tables and text for question answering, 
% making it a suitable dataset to evaluate the retriever performance.
% Considering the scale of the dataset and the evidence-aggregation characteristic of questions, we believe OTT-QA is an extremely suitable dataset to evaluate our joint retriever. 
Data statistics of OTT-QA dataset and table-text corpus are shown in Table \ref{tab:data-statistic}.

\begin{table}[h]
\centering
\resizebox{0.45\textwidth}{!}{
\begin{tabular}{ll}
\hline
% \# questions (train)     & 41,469  \\ 
% \# questions (dev)  & 2,214          \\
% \# questions (test)     & 2,158     \\
\# questions (train/dev/test)     & 41,469 / 2,214 / 2,158   \\ 
\# of tables in the corpus  &  410,740         \\
\# of passages in the corpus & 6,342,314 \\ \hline
\# of fused table-text blocks in the corpus  &  5,409,903         \\
Average tokens in fused blocks  &   357.5         \\ 
Average fused table-text blocks in each table  &   12.9         \\
% Average \# of tokens in fused blocks  &   357.5         \\ 
% Average \# of rows in each table  &   12.9         \\
\hline
\end{tabular}
}
\caption{Statistics of OTT-QA and table-text corpus.}
\label{tab:data-statistic} 
% \vspace{-4mm}
\end{table}

% \paragraph{Metrics}
% \subsection{Metrics}
% On the table-and-text retrieval task, we use table recall at top $k$ ranks and block recall at top $k$ ranks, as described in \cref{sec:evaluation-protocal}.
% % Following previous work \cite{chen2020ottqa, Kostic2021triencoder}, we use 
% % table recall at top $k$ ranks (TR@k) to evaluate retriever performance on the table-and-text retrieval task.
% % A table is considered a correct match if any of the deriving fused blocks is retrieved. 
% % A table is considered a correct match if the answer resides in the table.
% % However, TR@k is an imperfect metric for table-and-text retrieval since the a table may be mistakenly counted even the fused block uncovers the correct answer. 
% % Therefore we propose a more challenging metric called block recall at top $k$ ranks (BR@k) where a fused block is considered a correct match when it meets two requirements. First, it comes from the ground truth table. Second, it contains the correct answer. 
% On the downstream QA task, we report the exact match (EM) and F1 score \cite{chen2020ottqa} to evaluate OpenQA system. 
\subsection{Evaluation Metrics} \label{sec:evaluation-protocal}
Both recall and accuracy can reflect retrieval performance and are commonly used in IR. In this paper, we use the recalls rather than precision-based metrics since they are widely used metrics in recent retrievers like MVR \cite{Zhang2022MVR} and RocketQA \cite{Qu2021rocketqa}.
Recall@$k$ aims to capture answers in top-$k$ results and ensure answers can be seen as the input of QA models, which is computed as the proportion of relevant items found in the top-$k$ returned items. 
% A well-recognized metric for information retrieval is the recall at top $k$ ranks (Recall@$k$), which is the proportion of relevant items found in the top-$k$ returned items. 

In this paper, we use two metrics to evaluate the retrieval system: one is table recall and the other is table-text block recall. 
% One metric is the table recall, where the gold table is the ground-truth table annotated by human. 
Table recall indicates whether the top-$k$ retrieved blocks come from the ground-truth table.
However, in table-and-text retrieval, table recall is imperfect as an coarse-grained metric since our basic retrieval unit is a table-text block corresponding to a specific row in the table. 
Therefore we propose a more fine-grained and challenging metric: table-text block recall at top $k$ ranks, where a fused
block is considered as a correct match when it meets two requirements. 
Firstly, it comes from the ground truth table. Second,
it contains the correct answer.
On the downstream QA task, we report the exact match (EM) and F1 score \cite{chen2020ottqa} to evaluate OpenQA system.
\section{Experiments: Table-and-Text Retrieval}
In this section, we evaluate the retrieval performance of our OpenQA Table-Text Retriever (\textsc{OTTeR}). We first compare \textsc{OTTeR} with previous retrieval approaches on OTT-QA. Then we conduct extensive experiments to examine the effectiveness of the three proposed mechanisms. 

\subsection{Experiment Settings}
% \vspace{-1mm}
% \subsection{Baseline Methods}
\paragraph{Baseline Methods}
We compare with the following 8 retrievers.
% \begin{itemize}
    % \item[(1)] 
    (1) \textbf{BM25 w/o text} \cite{chen2020ottqa} is a sparse method to retrieve tabular evidence, where the flat table with metadata (i.e., table title and section title) and content are used for retrieval. 
    % \item 
    % \item[(2)] 
    % (2) \textbf{Bi-Encoder} \cite{Kostic2021triencoder} is a dense retriever which uses a BERT encoder for questions, and a shared BERT encoder to separately encode tables and text as representations for retrieval.
    % % \item
    % % \item[(3)] 
    % (3) \textbf{Tri-Encoder} \cite{Kostic2021triencoder} is a dense retriever that uses three individual BERT encoders to separately encode questions, tables and text as representations.
    (2)  \textbf{Iterative Retriever} \cite{chen2020ottqa} is a dense retriever which iteratively retrieves tables and passages in 3 steps.
    (3) \textbf{Fusion Retriever} \cite{chen2020ottqa} is the \textit{only} existing dense method to retrieve table-text block, which uses a GPT2 \cite{radford2019gpt2} to link passages and the Inverse Cloze Task \cite{Lee2019LatentRF} to pre-train the encoder. % \end{itemize}
    We directly reports results of the above three models from original papers. As the only reported retrieval performance of Iterative/Fusion Retriever is Hit\@4K (whether the answer exists in the retrieved 4096 tokens), for fair comparison, we also report results of our implemented models with Hit\@4K. 
    % (4) \textbf{Bi-Encoder} \cite{Kostic2021triencoder} is a dense retriever which uses a BERT encoder for questions, and a shared BERT encoder to separately encode tables and text as representations for retrieval.
    % (5) \textbf{Tri-Encoder} \cite{Kostic2021triencoder} is a dense retriever that uses three individual BERT encoders to separately encode questions, tables and text as representations.
    (4) \textbf{Bi-Encoder} and (5) \textbf{Tri-Encoder} \cite{Kostic2021triencoder} are two dense retrievers which encode question, table and text with seperate BERT, where the former uses a shared encoder for table and text, and the latter uses three encoders for the input. As they don't release code and models, we report our reproduced results in our setting.  
As there are few works on table-text retrieval, we also implement the following baselines to retrieve from the same corpus in \textsc{OTTeR}. (6) \textbf{BM25} is a sparse method to retrieve table-text blocks. 
(7) \textbf{\textsc{OTTeR}-baseline} is a dense retriever for table-text blocks using random negatives without MER and pre-training.
(8) \textbf{\textsc{OTTeR} w/o text} is a dense retriever to retrieve table evidences. We remove textual passages in corpus during retrieval.
% (removing textual passages during retrieval).

% \vspace{-1mm}
% \subsection{Implementation Details}
\paragraph{Implementation Details}
% All of our code is based on PyTorch \cite{Paszke2019PyTorchAI} and HuggingFace's Transformers \cite{Wolf2019HuggingFacesTS}.  
We use RoBERTa-base \cite{Liu2019roberta} as the backbone of our retrievers since RoBERTa is an improved version of BERT \cite{Devlin2019BERT}. The retrievers take as the input a maximum input length of 512 tokens per table-text block and 70 tokens per question. The retrievers are trained using in-batch negatives and one additional hard negative for both pre-training and fine-tuning.
On the pre-training stage, we pre-train on the synthesized corpus for 5 epochs on 8 Nvidia Tesla V100 32GB GPUs with a batch size of 168.
% The dual-encoder is trained with the batch size of 168 and the hard negatives are randomly selected.
% We randomly select a hard negative and apply in-batch negatives during training. 
We use AdamW optimizer \cite{Loshchilov2019AdamW} with a learning rate of 3e-5, linear scheduling with 5\% warm-up.
On the fine-tuning stage, we train all retrievers for 20 epochs with a batch size of 64, learning rate of 2e-5 and warm-up ratio of 10 \% on 8 Nvidia Tesla V100 16GB GPUs. 
% In-batch negatives and hard negatives in \cref{sec:OTTER} are also used.
% Hard negatives are described in Section \ref{sec:MMHN}.
% We do not conduct hyperparameter search for both pre-training and fine-tuning tasks.

\begin{table}[tbp]
% \small
  \centering
  \resizebox{0.48\textwidth}{!}{
    \begin{tabular}{lccccc|c}
    \hline
    Models & R@1  & R@10 & R@20 & R@50 & R@100 &Hit@4K \\
    \hline
    % BM25 \cite{chen2020ottqa} & 41.0  & 68.5  & 73.7  & 80.4  & - \\
    % Bi-Encoder \cite{Kostic2021triencoder} & -     & 72.9  & 78.0  & -     & 89.4 \\
    % Tri-Encoder \cite{Kostic2021triencoder} & -     & 73.8  & 79.7  & -     & 90.1 \\
    BM25 w/o text & 41.0  & 68.5  & 73.7  & 80.4  & - & - \\
    % Bi-Encoder (Kosti'c et al. 2021) & -     & 72.9  & 78.0  & -     & 89.4 & - \\
    % Tri-Encoder (Kosti'c et al. 2021) & -     & 73.8  & 79.7  & -     & 90.1 & - \\
    Bi-Encoder* & 46.2     & 70.9  & 76.0  & 82.1  & 88.4 & 44.2 \\
    Tri-Encoder* & 47.7    & 70.8  & 77.7  &  82.5  & 88.1 & 46.3 \\
    Iterative Retriever  & - & -  & -  & -  & - & 27.2 \\
    Fusion Retriever     & - & -  & -  & -  & - & 52.4 \\
    %  BM25 & 41.0  & 68.5  & 73.7  & 80.4  & - \\
    % % Fusion Retriever &       &       &       &       &  \\
    % Bi-Encoder & -     & 72.9  & 78.0  & -     & 89.4 \\
    % Tri-Encoder & -     & 73.8  & 79.7  & -     & 90.1 \\
    \hline
    BM25 & 44.5  & 69.1  & 74.2  & 80.2  & 89.8  & 48.0 \\
    \textsc{OTTeR}-baseline  & 46.3  &  69.4  &  74.4  &  80.1 &  83.9 &  54.6 \\ 
    % \textsc{OTTeR} w/o MER \& MMHN \& MSP  & 46.3  &  69.4  &  74.4  &  80.1 &  83.9 \\ 
    \textsc{OTTeR} w/o text & 48.7    & 73.9  &  79.5  & 85.8     &  88.8  & 34.6 \\ 
    \textsc{OTTeR} & \textbf{58.5} & \textbf{82.0} & \textbf{86.3} & \textbf{90.6} & \textbf{92.8}  & \textbf{66.4}\\
    \hline
    \end{tabular}%
    }
    
  \caption{Overall retrieval results on OTT-QA dev set. Table recall and Hit@4K \cite{chen2020ottqa} are reported, where Hit@4K is used to measure whether the answer exists in the retrieved 4096 subword tokens. * denotes the results reproduced by us. }
%   \caption{Overall results of table retrieval on OT/T-QA dev set. Table recalls are reported. We also report results of  \textsc{OTTeR}-baseline (removing three proposed strategies) and \textsc{OTTeR} w/o text (removing textual passages during retrieval).}
%   \caption{Overall results of table retrieval on OTT-QA dev set. Table recalls are reported. We also compare \textsc{OTTeR} without the three proposed strategies (\textsc{OTTeR}-baseline) and textual  passages  are  removedduring retrieval (OTTER w/o text),}
  \label{tab:retrieval-overall}%
%   \vspace{-4.5mm}
\end{table}%

\subsection{Main Results}
Table \ref{tab:retrieval-overall} compares different retrievers on OTT-QA dev. set, using the table recall at top $k$ ranks ($k \in \{1, 10, 20, 50, 100\}$) because the results from other papers are mainly reported in table recall. 
% We also report the result after removing the three proposed strategies (\textsc{OTTeR}-baseline). 
% There are several findings:
We find that:
(1) \textsc{OTTeR} significantly outperforms previous sparse and dense retrievers and the gap is especially large when $k$ is smaller (e.g., 8.2\% absolute gain for R@10), which demonstrates the effectiveness of \textsc{OTTeR};
% (2) Comparing with Bi-Encoder and Tri-Encoder which separately retrieve tables and text, \textsc{OTTeR} produces better results, which is presumably because tables and text are fused
% (3) We also report the performance of \textsc{OTTeR} when only tabular information is given during retrieval (\textsc{OTTeR} w/o text). 
(2) When textual passages are removed during retrieval (\textsc{OTTeR} w/o text), the performance of \textsc{OTTeR} drops dramatically, especially when $k$ is smaller. 
This phenomenon shows the importance of taking textual information as a complement to tables.

\subsection{Ablation Study}\label{sec:result-ablation}
To examine the effectiveness of the three mechanisms in \textsc{OTTeR}, we conduct extensive ablation studies on OTT-QA and discuss our findings below.

% In this part, we conduct the extensive experiments on OTT-QA dataset to examine the effectiveness of the three strategies  
\subsubsection*{Effect of Modality-enhanced Representation}

In this experiment, we explore the effect of modality-enhanced representations (MER) on retrieval performance. 
Table \ref{tab:modality-enhanced-repr} reports the table recall and block recall of our models with different MER strategies on the OTT-QA dev. set. 
We also report the result after eliminating MER, i.e., using only the representation of the \texttt{[CLS]} token for ranking.
% ablation results and the analysis plot figure
We find that integrating modality-enhanced representations improves the retrieval performance significantly.
% , showing the benefits of enriched representations of 
% is extremely beneficial to table-and-text retrieval. 
% As MER provides extra unique representations for each single modality, retrievers can easily capture the comprehensive semantics of fused blocks. 
As MER incorporates single-modal representations to enrich the mixed-modal representation, retrievers can easily capture the comprehensive semantics of table-text blocks. 
In addition, among all the strategies for MER, the FIRST strategy using the representation of the beginning special token of each modality achieves the best performance. This observation verifies the stronger representative ability of the FIRST strategy compared with other pooling strategies. 
% the MER and using the embeddings of special tokens as enhanced representations, i.e., MER=FIRST, obtains the best result, which can demonstrate the stronger representation ability of special tokens in dense retrieval than other pooling strategies to some extend. 
% , which is consistent with the findings in \citet{}.

\begin{table}[t]
% \small
  \centering
  \resizebox{0.43\textwidth}{!}{
    \begin{tabular}{l|ccc|ccc}
    \hline
    & \multicolumn{3}{c|}{Table Recall} & \multicolumn{3}{c}{Block Recall} \\
    % \hline
    Models & R@1  & R@10 & R@100 & R@1  & R@10 & R@100 \\
    \hline
    \textsc{OTTeR} &  &  &  &  &  & \\
    \quad MER=FIRST  & \textbf{58.5} & \textbf{82.0} & \textbf{92.8} & \textbf{30.9} & \textbf{66.4} & \textbf{87.0} \\
    \quad MER=AVG  & 57.1  & 81.2  & 92.5  & 29.8  & 65.3  & 85.9 \\
    \quad MER=MAX  & 56.7  & 81.4  & 92.2  & 29.0  & 65.1  & 86.4 \\
    \quad MER=SelfAtt  & 57.9  & 81.2  & 92.6  & 29.5  & 65.3  & 86.0 \\
    \quad w/o MER & 50.0  & 76.8  & 89.9  & 22.7  & 55.2  & 79.3 \\
    \hline
    \end{tabular}
    }
  \caption{Retrieval performance of \textsc{OTTeR} under different modality-enhanced representations (MER) settings.}
  \label{tab:modality-enhanced-repr}
%   \vspace{-3mm}
\end{table}%

\subsubsection*{Effect of Mixed-modality Negative Sampling}
To investigate the effectiveness of hard negative sampling on retrieval, we evaluate our system 
% with following hard negative sampling methods on the OTT-QA development set.
under following settings of hard negative sampling on the OTT-QA development set:
% of models trained with different hard negatives, using the following settings: 
(1) Mixed-modality hard negative (MMHN) described in \cref{sec:MMHN}; 
(2) BM25: the most similar irrelevant table-text block searched by BM25;
(3) Random: a random table-text block in the same table containing no answer.

From the results shown in Table \ref{tab:hard-negative}, we can observe that training the retriever with MMHN yields the best performance compared with other hard negative sampling strategies.
% trained with MMHN, the retriever yields higher recalls than simply using BM25 or Random hard negatives. 
Since mixed-modality hard negatives is constructed by only replacing partial information from the positive block, it is more challenging and it enables the retriever to better distinguish important information in the evidence.
% cover more similar information with the positive block, they are potentially harder than BM25 and Random, which presumably improve the discriminant ability of models and thus derive better representations.
% In this experiment, we investigate 

\begin{table}[tbp]
% \small
  \centering
  \resizebox{0.43\textwidth}{!}{
    % \begin{tabular}{l|ccc|ccc}
    \begin{tabular}{l|ccc|ccc}
    \hline
    & \multicolumn{3}{c|}{Table Recall} & \multicolumn{3}{c}{Block Recall} \\
    % \hline
    Models & R@1  & R@10 & R@100 & R@1  & R@10 & R@100 \\
    \hline
    \textsc{OTTeR} &  &  &  &  &  & \\
    \quad HN=MMHN  & \textbf{58.5} & \textbf{82.0} & \textbf{92.8} & \textbf{30.9} & \textbf{66.4} & \textbf{87.0} \\
    \quad HN=BM25 & 51.4  & 79.8  & 92.4  & 25.8  & 58.2  & 81.9 \\
    % \quad HN=BM25 (out table) & 54.3  & 77.0  & 90.6  & 23.8  & 58.5  & 82.4 \\
    \quad HN=Random & 50.3  & 79.0    & 92.7  & 28.4  & 58.7  & 80.1 \\
    % \quad HN=Random (out table) & 57.8  & 78.4    & 91.2  & 30.0  & 67.2  & 88.6 \\
    \hline
    \end{tabular}
    }
  \caption{Retrieval performance of \textsc{OTTeR} under different hard negative sampling settings. MMHN denotes mixed-modality hard negatives. }
  \label{tab:hard-negative}
%   \vspace{-4mm}
\end{table}%

\subsubsection*{Effect of Mixed-modality Synthetic Pre-training}
We investigate the effectiveness of mixed-modality synthetic pre-training.
We first pre-train the retriever and then fine-tune the retriever with OTT-QA training set.
The pre-training corpus consisting of 3 millions of \textit{(question, evidence)} pairs, 
with questions synthesized in the following ways:
(1) \textbf{BartQ}: the questions are generated by BART as described in \cref{sec:PT}; 
% synthesized table-text questions (STQ), where questions are generated by BART as described in \cref{sec:PT};
(2) \textbf{TitleQ}: the questions are constructed from passage titles and table titles.
(3) \textbf{DA w/o PT}: data augmentation without pre-training, where we integrate the BART synthetic corpus with the oracle data together for fine-tuning. 
(4)\textbf{w/o PT} direct fine-tuning without pre-training. 
% (3) w/o PT: directly fine-tuning on OTT-QA without pre-training (PT);
% (4) DA: data augmentation (DA) with synthesized questions, where we add the BART generated pairs into training set as augmented data and then fine-tuning on the merged training set.

The retrieval results on the dev. set of OTT-QA are exhibited in Table \ref{tab:pretrain}. 
We can find that:
(1) Pre-training brings substantial performance gain to dense retrieval, showing the benefits of automatically synthesizing large-scale pre-training corpus to improve retrievers.
% A major merit of pre-training is that it does not explicitly rely on manually-labeled data. Instead, it utilizes human heuristics or generators to create large-scale synthesized corpus to improve the dual-encoder. 
(2) synthesizing questions using BART-based generator performs better than using template-based method (TitleQ).
We attribute it to more fluent and diverse questions synthesized by generation-based method.
% because generation-based method can synthesize more fluent and diverse questions;
% comparing two pre-training corpus (PT=STQ and PT=Title), we find pre-training with STQ performs better. 
% We attribute it to the higher quality of pre-training data since BART can generate more fluent evidence-gathering questions than simply using titles. 
(3) Using the synthesized corpus for data augmentation performs much poorer than using it for pre-training, and even worse than directly fine-tuning without pre-training.
One explanation is that pre-training targets to help the model in learning a more general retrieving ability beforehand, while fine-tuning aims to learns a more specific and accurate retriever.
As the synthesized corpus is more noisy, using it as augmented fine-tuning data may make the training unstable and lead to a performance drop.
% Therefore, fine-tuning with the noisy synthesized corpus could  noise to the final retriever.
This observation again verifies the effectiveness of pre-training with mixed-modality synthetic corpus. 
% Using it for pre-training could help
% Compared with 41K high-quality training examples, 3M synthesized corpus bear relatively low quality and make the training unstable.
% Therefore, the extra synthesized corpus 
% One possible reason is that extra synthesized corpus leads to more noise during fine-tuning, while pre-training is targeted to 
% enables the model to learn retrieval ability beforehand.
% That is probably because the average length of the documents is much longer than the length of passages and our method can make full use of aggregating the semantics of the whole document.

\begin{table}[t]
% \small
  \centering
  \resizebox{0.475\textwidth}{!}{
    \begin{tabular}{l|ccc|ccc}
    \hline
    & \multicolumn{3}{c|}{Table Recall} & \multicolumn{3}{c}{Block Recall} \\
    % \hline
    Models & R@1  & R@10 & R@100 & R@1  & R@10 & R@100 \\
    \hline
    \textsc{OTTeR} &  &  &  &  &  & \\
    \quad  PT=BartQ & \textbf{58.5} & \textbf{82.0} & \textbf{92.8} & \textbf{30.9} & \textbf{66.4} & \textbf{87.0} \\
    \quad  PT=TitleQ & 56.6  & 79.3  & 91.8  & 23.1  & 60.0  & 83.1 \\
    % \hline
    \quad  DA w/o PT & 39.3  & 68.9  & 73.0    & 14.8  & 45.9  & 74.5 \\
    \quad  w/o PT & 53.1  & 77.8  & 91.2  & 20.5  & 57.2  & 81.3 \\

    \hline
    \end{tabular}
    }
  \caption{Retrieval performance of \textsc{OTTeR} under different settings. PT denotes pre-training.}
  \label{tab:pretrain}
%   \vspace{-3mm}
\end{table}%

\begin{figure*}[thbp]
     \centering
     \includegraphics[width=0.95\textwidth]{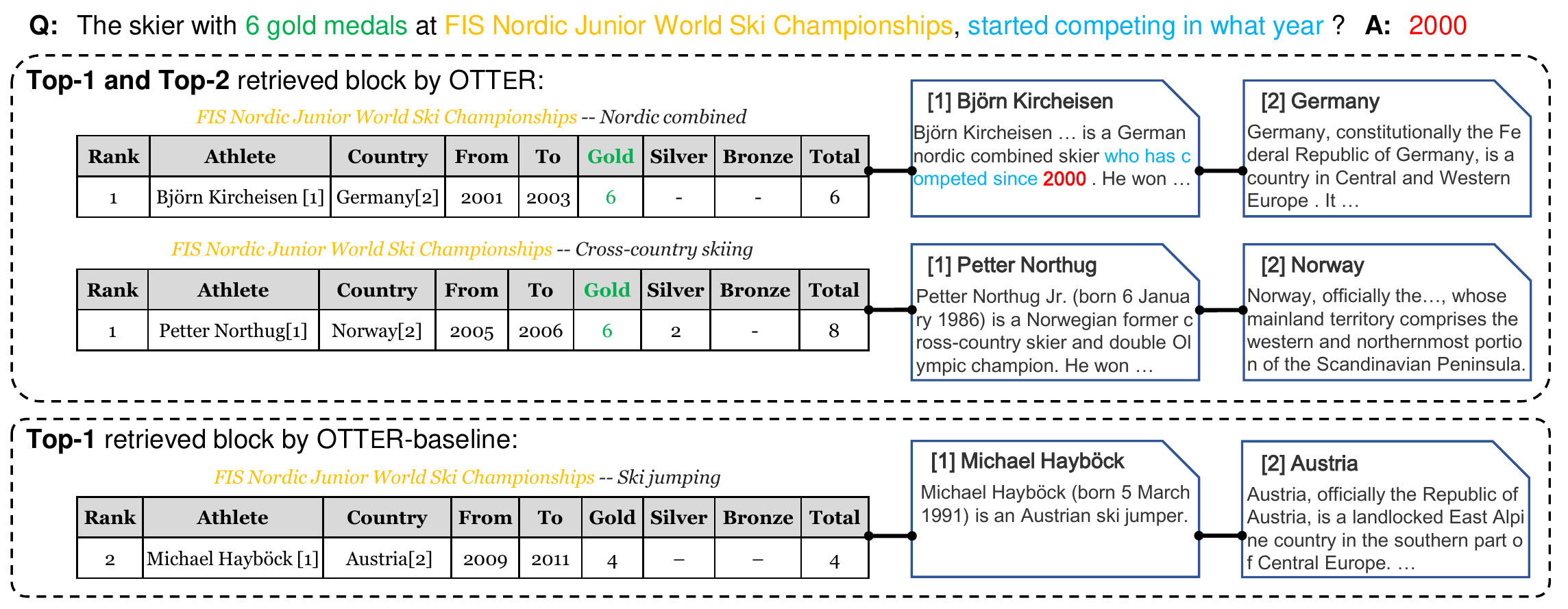}
    %  \vspace{-1mm}
     \caption{Examples of table-text blocks returned by full \textsc{OTTeR} and \textsc{OTTeR} without modality-enhanced representations. Words in the retrieved blocks of the same color denote the evidences corresponding to questions.}
     \label{fig:case-study}
    %  \vspace{-3mm}
\end{figure*}

\subsection{Case Study}
Here, we give an example of retrieved evidences to show that \textsc{OTTeR} correctly represents questions and blocks with the proposed three strategies.
% modality-enhanced representations (MER). 

As shown in Figure \ref{fig:case-study}, to answer the question, the model should find relevant table-text blocks with two pieces of evidences distributed in tables and passages, including the 
% \textit{``skier who won 6 gold medals at FIS Nordic Junior World Ski Championships"} and the \textit{``year when the skier started competing"}. 
\textit{``skier who won 6 gold medals at the FIS Nordic Junior World Ski Championships"} and the \textit{``year when the skier started competing"}. 
% The two evidences are separated in the heterogeneous sources, i.e. tables and passages. 
As we can see, 
% incorporated with three mechanisms, 
\textsc{OTTeR} successfully returns a correct table-text block at rank 1, which includes all necessary information. 
The top-2 retrieved block by \textsc{OTTeR} is also reasonable, since partial evidences like \textit{6 gold medals} and 
% \textit{FIS Nordic Junior World Ski Championships} 
\textit{Ski Championships} are matched. 
% However, the top-1 retrieved block by \textsc{OTTeR} without MER is an unsatisfactory hit. 
However, \textsc{OTTeR}-baseline (w/o three mechanisms) returns an unsatisfactory block.
Though the retriever finds 
% important information of 
% \textit{FIS Nordic Junior World Ski Championships}
the \textit{Ski Championships}
, which is a strong signal to locate the table, it fails to capture fine-grained information like \textit{6 gold medals} and \textit{starting year}.

This case demonstrates that \textsc{OTTeR} can capture the more accurate meanings of fused table-text block, especially when the supported information resides separately. 
% It also shows that enhancing representations by explicitly incorporating single modality is beneficial to modeling bi-modal data.
It shows that enhancing cross-modal representations with proposed mechanisms is beneficial to modeling heterogeneous data.
\section{Experiments: Question Answering}
% \vspace{-1mm}
In this section, we experiment  to show how \textsc{OTTeR} affects the downstream QA performance. 

% \vspace{-2mm}
\subsection{Reader} 
We implement a two-stage open-domain question answering system, which is equipped with our \textsc{OTTeR} as the \textit{retriever} and a \textit{reader} model for extracting the answer from the retrieved evidence.
% ,where we can plug in our retriever \textsc{OTTeR} directly. 
% Besides the \textit{retriever}, our OpenQA system is equipped with a neural \textit{reader} that extracts answers given the question and evidences.
As we mainly focus on improving the retriever in this paper, we use the state-of-the-art reader model to evaluate the downstream QA performance. 

Following \citet{chen2020ottqa}, we use the \textit{Cross Block Reader} (CBR) to extract answers. 
The CBR jointly reads the concatenated top-$k$ retrieved table-text blocks  and outputs a best answer span from these blocks. 
In contrast to \textit{Single Block Readers} (SBR) that read only one block at a time, CBR is more powerful in utilizing the cross-attention mechanism to model the cross-block dependencies. 
We take the pre-trained Long-Document Transformer (Longformer) \cite{beltagy2020longformer} as the backbone of CBR, which applies sparse attention mechanism and accepts longer input sequence of up to 4,096 tokens. 
For fair comparison with \citet{chen2020ottqa}, we feed top-$15$ retrieved blocks into the reader model for inference.
% During training, we first fine-tune CBR on SQuAD 2.0 and then on OTT-QA. 
To balance the distribution of training data and inference data, we also takes $k$ table-text blocks for training, which contains several ground-truth blocks and the rest of retrieved blocks.  
% For each instance, we sample 14 more hard negative blocks from the retriever for each question, and mix them with the ground truth fused block to train. 
The training objective is to maximize the marginal log-likelihood of all the correct answer spans in the positive block. 
The reader is trained with 8 Nvidia V100 GPUs for 5 epochs with a batch size of 16 and learning rate of 1e-5.

\begin{table}[tbp]
%   \vspace{2mm}
% \small
  \centering
    \resizebox{0.475\textwidth}{!}{
    \begin{tabular}{lcccc}
    \hline
        & \multicolumn{2}{c}{Dev} & \multicolumn{2}{c}{Test} \\
    \hline
    Retriever + Reader & EM    & F1    & EM    & F1 \\
    \hline
    BM25 + HYBRIDER \cite{chen2020hybridqa}  & 10.3  & 13.0  & 9.7   & 12.8 \\
    BM25 + DUREPA \cite{li2021durepa}   & 15.8  & -     & -     & - \\
    Iterative Retriever + SBR \cite{chen2020ottqa}  & 7.9   & 11.1  & 9.6   & 13.1 \\
    Fusion Retriever + SBR \cite{chen2020ottqa}  & 13.8  & 17.2  & 13.4  & 16.9 \\
    Iterative Retriever + CBR \cite{chen2020ottqa}  & 14.4  & 18.5  & 16.9  & 20.9 \\
    Fusion Retriever + CBR \cite{chen2020ottqa}  & 28.1  & 32.5  & 27.2  & 31.5 \\
    \hline
    \textsc{OTTeR}-baseline +CBR & 27.1 & 32.5 & - & - \\
    \textsc{OTTeR} (w/o MER) + CBR & 33.7 & 39.4 & - & - \\
    \textsc{OTTeR} (HN=random) + CBR & 34.6 & 40.5 & - & - \\
    \textsc{OTTeR} (w/o PT) + CBR & 32.4 & 38.7 & - & - \\
    \textsc{OTTeR}  + CBR & \textbf{37.1} & \textbf{42.8} & \textbf{37.3} & \textbf{43.1} \\
    \hline
    \end{tabular}
    }
  \caption{QA Results on the dev. set and blind test set.}
  \label{tab:openqa}
%   \vspace{-4mm}
\end{table}

% To improve we first fine-tuned Longformer on SQuAD 2.0 \cite{squad} 
% We use the Exact Match (EM) and F1 score to measure the performance of the reader.

% \subsection{Model Comparison} 
% % \subsubsection*{Baselines} 
% \paragraph{Baselines}
% We compare our system with methods below: 
% % \begin{itemize}
%     % \item[(1)] 
%     (1) \textbf{BM25+HYBRIDER} \cite{chen2020hybridqa} is a system that first adopts BM25 to retrieve tables and passages separately and then uses a two-stage model to perform multi-hop reasoning. 
%     % \item[(2)]  
%     (2) \textbf{Iterative Retriever (IR) / Fusion Retriever (FR) + Single Block Reader (SBR) / Cross Block Reader (CBR)} are proposed by \citet{chen2020ottqa}. IR and FR are dense retrievers where IR separately retrieves tables and passages and FR retrieves them as a table-text block. SBR is a standard reader that reads a question and a block at a time while CBR reads the top-k blocks to output the answer. 
%     % \item[(3)] 
%     (3) \textbf{BM25+DUREPA} \cite{li2021durepa}  firsts adopts BM25 to retrieve tables and passages separately and then uses a dual reading-parsing model to output answers.
% % \end{itemize}

% \vspace{-3mm}
\begin{figure}[t]
     \centering
     \includegraphics[width=0.475\textwidth]{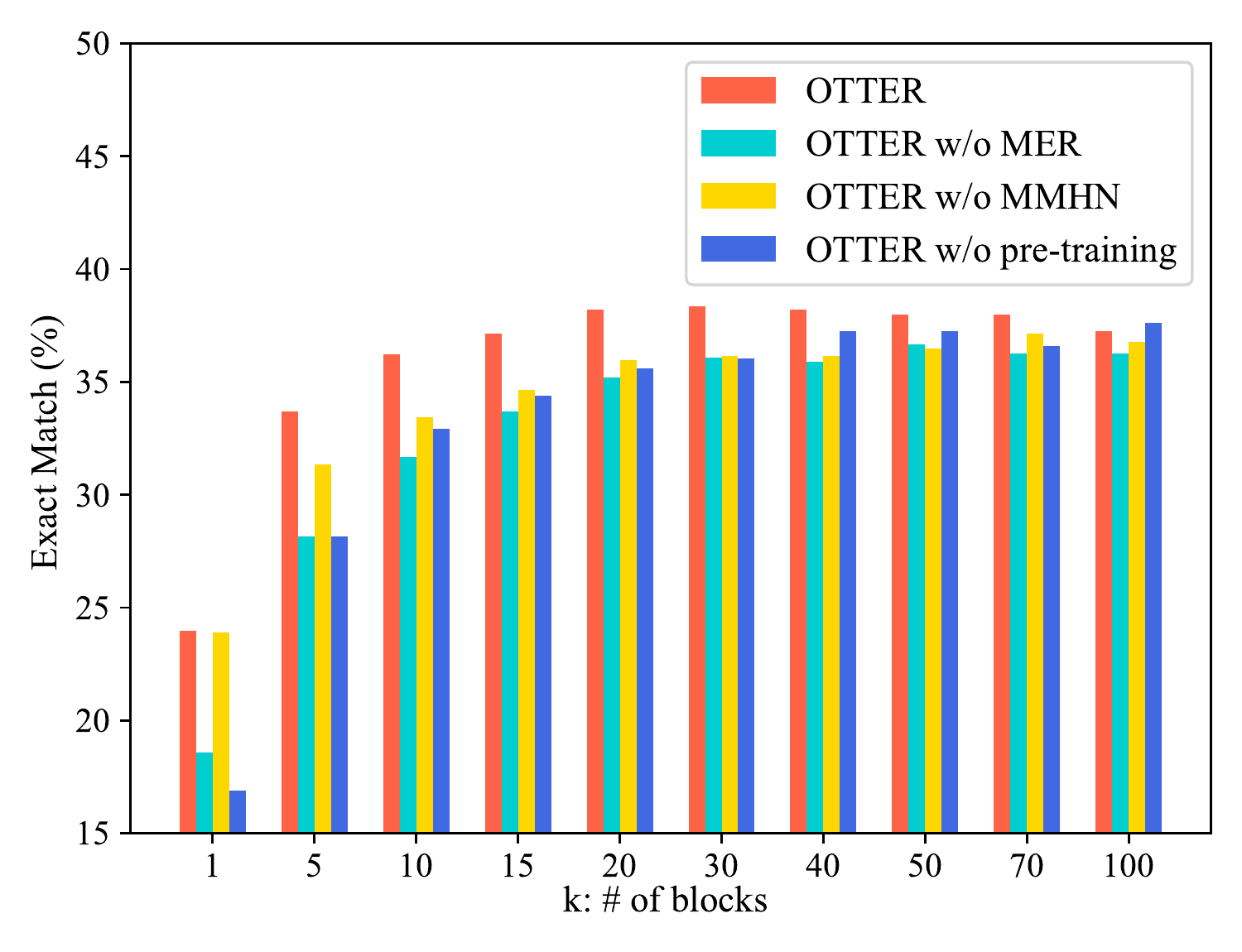}
    %  \vspace{-2mm}
     \caption{QA results of EM on the OTT-QA dev. set with different number of table-text blocks as input. }
     \label{fig:qa-result}
    %  \vspace{-5mm}
\end{figure}

% \subsubsection{Results}
% \paragraph{Results}
% \vspace{0.5mm}
\subsection{Results}
The results are shown in Table \ref{tab:openqa}. We find that \textsc{OTTeR} + CBR significantly outperforms existing OpenQA systems, with 10.1\% performance gain from test set in terms of EM over the prior state-of-the-art system and 10.0 \% EM gain from dev set over \textsc{OTTeR}-baseline + CBR.
% As our contributions lie in the dense retrieval component, t
The results demonstrate that our proposed three approaches can retrieve better supported evidences to the question, which leads to further improvement on the downstream QA performance. 
% Comparing the QA results with 

% We also report QA results with different number of input fused blocks to CBR. As shown in Figure \ref{fig:qa-result}, the EM score increases rapidly with $k$ when $k < 15$ but the growth slows down and even turns negative when $k > 15$. This demonstrates the 
To further analyze the effect of different components of \textsc{OTTeR} on QA performance, we conduct an ablation study on OTT-QA after eliminating different components.
As shown in Figure \ref{fig:qa-result}, 
the OpenQA system with full \textsc{OTTeR} achieves the best performance, and removing each component leads to a substantial performance drop.
% outperforms those with other \textsc{OTTeR} components. 
This observation verifies the effectiveness of our proposed three mechanisms, i.e., modality-enhanced representations (MER), mixed-modality hard negatives (MMHN) and mixed-modality synthetic pre-training. 
% As shown in Figure \ref{fig:qa-result}, the EM score increases rapidly with $k$ when $k < 15$ but the growth slows down and even turns negative when $k > 15$. This demonstrates the 
We also evaluate the impact of taking different numbers of retrieved blocks as the inputs for inference. As shown in Figure \ref{fig:qa-result}, the EM score increases rapidly with $k$ when $k < 20$ but the growth slows down when $k > 20$, which can help to find a better tradeoff between efficiency and performance.

% \vspace{-1mm}
\section{Related Works}
% \vspace{-1mm}
% dense retrieval
In OpenQA \cite{Chen2017ReadingWT,Joshi2017TriviaQA,  dunn2017searchqa, Lee2019LatentRF}, the retriever is an essential component to identify relevant evidences for answer extraction. 
In contrast to sparse information retrieval methods \cite{Wang2018R3, Nogueira2019PassageBM25, Yang2019BERTserini}, recent OpenQA systems tend to adopt dense retrieval approaches utilizing dense representations learned by pre-trained language models \cite{Lee2019LatentRF, Guu2020realm, karpukhin2020dense}. 
These methods are powerful in capturing contextual semantics. 

% which capture better semantic matching and resolve the problem of term mismatch. 
The prevailing OpenQA datasets mainly take the unstructured passage as evidence, including Natural Questions \cite{Kwiatkowski2019NaturalQuestions},  TriviaQA \cite{Joshi2017TriviaQA}, WebQuestions \cite{Berant2013WebQuestions}, CuratedTREC \cite{Baudis2015CuratedTrec} and SQuAD \cite{Rajpurkar2016SQuAD1Q}. 
There are also recent works studying OpenQA in the tabular domain \cite{Herzig2021nqtables, zhu2021TATQA, Zhong2022CARP}. \citet{chen2020ottqa} consider a more challenging setting that takes both tabular corpus and textual corpus as the knowledge sources, which is also the setting in this paper.
% differences to existing works

Our approach differs from existing methods mainly in two aspects: targeted evidence source and mixed-modality learning mechanisms.
First of all, we retrieve mixed-modality evidence from both tabular and textual corpus, which is different from text-based retrievers \cite{karpukhin2020dense,Asai2020LearningTR, Xiong2021mdr, xu2022LaPraDoR} and table-based retrievers \cite{Chen2020TableSU, Shraga2020WebTR, Pan2021CLTR}.
% (1) targeted evidence sources. Compared to textual OpenQA retrievers \cite{karpukhin2020dense,Asai2020LearningTR, Xiong2021mdr} and table retrievers \cite{Chen2020TableSU, Shraga2020WebTR, Pan2021CLTR} , we explore a more realistic and challenging task where the bi-modal information is retrieved. 
Secondly, our proposed three mixed-modality learning mechanisms also differ from existing methods. 
As for mixed-modality representation, previous work \cite{karpukhin2020dense} mainly uses the single representation of the special token for ranking. Our method incorporates single modal representation to enrich the mixed modal representation.  
As for mixed-modality negative sampling, instead of finding an entire negative evidence with either sparse or dense methods \cite{Yang2021xMoCoCM, Luan2021SparseDA,Lu2020NegativeContrast, Xiong2021ance, Lu2021LessIM, Zhan2021OptimizingDR, zhang2022ar2, xiao2022distillvq}, we construct more challenging hard negative by only replacing partial single-modality information at once. 
As for mixed-modality synthetic pre-training, our pre-training strategy is different in the pre-training task, knowledge source and the method of synthesizing pseudo question. 
There are also works investigating joint pre-training over tables and text \cite{Herzig2020tapas, Eisenschlos2020UnderstandingTW,Yin2020TaBERT, Ouz2020UniKQA}. However, these methods mainly take the table metadata as the source of text and do not consider the retrieval task. Instead, we use linked passages as a more reliable knowledge source, and target on retrieval-based pre-training.
% But they do not consider the retrieval task.
% (2) Method. Our proposed three training strategies also differ from existing methods. 
% Recent attempts mine hard negatives with sparse BM25 \cite{, Yang2021xMoCoCM, Luan2021SparseDA} and dense neural networks \cite{Lu2020NegativeContrast, Xiong2021ance, Lu2021LessIM, Zhan2021OptimizingDR} to a retriever. Whereas we propose a simple but effective way to obtain hard negative with human heuristics. 
There are some attempts on incorporating pre-training task to improve retrieval performance \cite{Chang2020PretrainingRetrieval, Sachan2021EndtoEndTO,  Ouguz2021DomainmatchedPT, Wu2022UnsupervisedCA}, which target on textual-domain retrieval or using template-based method for query construction. 
Differently, our approach focuses on a more challenging setting that retrieves evidence from tabular and textual corpus and adopts a  generation-based query synthetic method. 
% with large-scale synthesized corpus is another effective way to boost retrieval accuracy \cite{Chang2020PretrainingRetrieval, Sachan2021EndtoEndTO,  Ouguz2021DomainmatchedPT}. But they focus on single-hop questions while we deal with a more complex scenario and naturally utilize table structure to synthesize.
% \citet{chen2020ottqa} and \citet{Kostic2021triencoder} also investigate table and text retrieval, but they only use the \texttt{[CLS]} representations which are insufficient to represent bi-modal data. 
Besides, \citet{Pan2021UnsupervisedMQ} explore to generate multi-hop questions for tables and text, but they focus on an unsupervised manner.

% Another two related works include 

% \vspace{-1.2mm}
\section{Conclusion}
% \vspace{-0.3mm}
% In this paper, we propose an optimized dense table-text retriever for OpenQA called \textsc{OTTeR} to address the table-text discrepancy and data sparsity problems.
In this paper, we propose an optimized dense retriever called \textsc{OTTeR}, to retrieve joint table-text evidences for OpenQA. 
\textsc{OTTeR} involves three novel mechanisms to address table-text discrepancy and data sparsity challenges, i.e., modality-enhanced representations, mixed-modality hard negative sampling, and mixed-modality synthetic pre-training. 
% In this paper, we propose a dense retriever called \textsc{OTTeR}, to jointly retrieve both tabular and textual evidences for OpenQA. 
% \textsc{OTTeR} involves three novel mechanisms to enhance mixed-modality representation learning, including modality-enhanced representations, mixed-modality hard negative sampling, and mixed-modality synthetic pre-training. 
We experiment on OTT-QA dataset and evaluate 
% the performance
on two subtasks, including retrieval and QA.
Results show that \textsc{OTTeR} significantly outperforms other retrieval methods by a large margin, which further leads to a substantial absolute performance gain of 10.1\% EM on the downstream QA. 
Extensive experiments illustrate the effectiveness of all three mechanisms in improving retrieval and QA performance. 
Further analyses also show the ability of \textsc{OTTeR} in retrieving more relevant evidences from heterogeneous knowledge resources.
% dense table-text retriever to improve the retrieval accuracy on the open table-text question answering task.
% We proposed three strategies in \textsc{OTTeR}, including enhancing representations of single modality, creating hard negatives by mixing modalities, and pre-training on synthesized corpus. 
% We proposed three strategies in \textsc{OTTeR}, including modality-enhanced representations, mixed-modality negative sampling, and pre-training with synthesized table-text questions.
% % We showed that the retrieval performance can be substantially boosted with the three critical ingredients. 
% Extensive experiments and analyses further confirm the effectiveness of the proposed approaches. 
% As a result of improved retrieval performance, we achieved new state-of-the-art results on OTT-QA benchmark, surpassing previous best system by a large margin.
% % We also demonstrate that the performance of end-to-end QA can be improved based on our \textsc{OTTeR} retriever.
% In the future, we can explore to extend the retriever to more modalities, e.g., images, etc.

\section*{Limitations}
Firstly, the mixed modality input bears a strong problem of input length. As the table size could be large, the size of linked passages could also be large, which leads to exceeding the maximum input length of PLMs. Thus we make a trade-off option to break down the table into rows, which fails to answer questions that require information among multiple rows. 
Secondly, OTT-QA is a fairly limited dataset, where answering each question MUST require  information from two modalities. 
% Thus it's not clear of the model's performance on a more real setting that we don't know the 
Thus it's not clear of the model's performance on single modal OpenQA where any one modality may not be required. It's also unclear whether incorporating tabular information will help textual OpenQA.
Thirdly, as the size of in-batch negative examples could heavily influence the retrieval performance, training a strong dense retriever requires large GPU resources, where we use 8 Nvidia Tesla V100 16GB GPUs for each experiment.

\section*{Acknowledgement}
We thank all anonymous reviewers for their valuable suggestions. We also thank Shunyu Zhang for fruitful discussions.

\bibliography{anthology,custom}
\bibliographystyle{acl_natbib}

% \newpage

\appendix

% \clearpage

\section{Method Details}
\subsection{Table-Text Block Representation}\label{app:block-representatoin}
\begin{figure}[ht]
     \centering
     \includegraphics[width=0.475\textwidth]{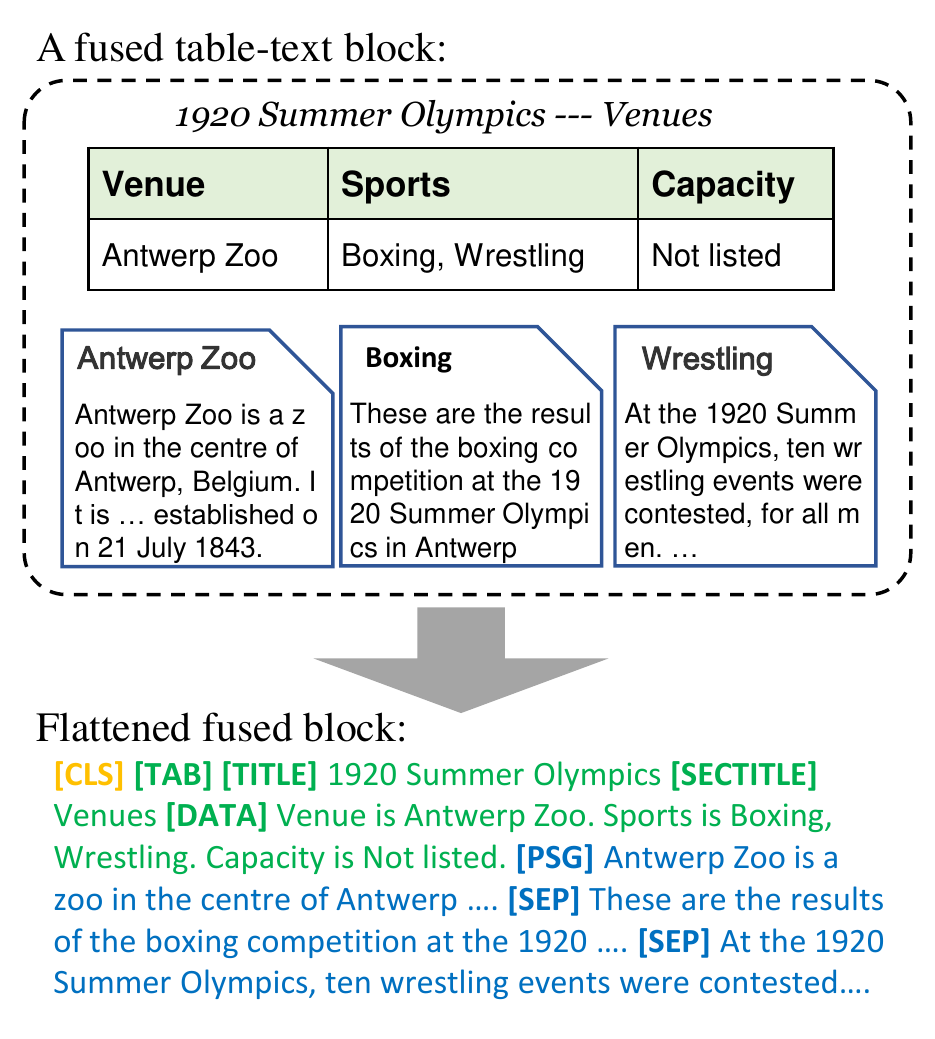}
     \caption{The flattened fused block representation of the each table-text block.}
     \label{fig:block-repre}
\end{figure}

The table-text block representation is illustrated in Figure \ref{fig:block-repre}. 
Following \citet{chen2020ottqa}, we involve the title and section title of a table and prefix them to the table cell. 
We also flatten the column name and column value with an ``is " token to obtain more natural and fluent utterance. 
In addition, we add different special tokens to separate different segments, including \texttt{[TAB]} for table segment,  \texttt{[PSG]} for passage segment, \texttt{[TITLE]} for table title, \texttt{[SECTITLE]} for section title, \texttt{[DATA]} for table content, and \texttt{[SEP]} to separate different passages. 
Such a flattened block will be used throughout this paper as the input string to the retriever and the reader.

In OTT-QA dataset, long rows frequently appear in tables, which leads to more entities and passages in a single table-text block. 
To maintain more relevant information in a block, we rank the passages with the TF-IDF score to table schema and table content.
Then we remove the tokens when a flattened block is out of the input length limit of the RoBERTa tokenizer.

% \subsection{Entity Linking}

\subsection{Examples of Synthesized Corpus}\label{app:synthetic-corpus}
To provide a better understanding of mixed-modality synthetics pre-training, we give some examples of pseudo training data with \textit{(question, table-text block)} pairs in Table \ref{tab:bart-example}. 
As we can see, the generated questions not only are fluent and natural, but also consider mixed-modality information from tables and passages.

\begin{table*}[htbp]
  \centering
  \small
%   \resizebox{1\textwidth}{!}{
    % \begin{tabular}{cc}
    % \begin{tabular}{p{33em}p{14.69em}}
    \begin{tabular}{p{33em}p{14em}}
    % \begin{tabular}{m{7.8cm}<{\centering}m{4.3cm}}
    \hline
    Flattened Table Segment & Generated Question \\
    \hline
    %  a & b \\
    \texttt{[TAB]} \texttt{[TITLE]} J1 League \texttt{[SECTITLE]} History -- Timeline \texttt{[DATA]} Year is 2003. Important events is Extra time. \# J clubs is 16. Rele . slots is 2. \texttt{[PSG]} The 2003 season was the 11th season since the establishment of the J.League . The league began on March 15 and ended on November 29 .
    & What is the number of slots in the J1 League for the season that began on March 15 and ended on November 29 ?
    \\
    \texttt{[TAB]} \texttt{[TITLE]} 2010 President's Cup (tennis) \texttt{[SECTITLE]} ATP entrants -- Seeds \texttt{[DATA]} Nationality is KAZ. Player is Mikhail Kukushkin. Ranking is 88. Seeding is 1. \texttt{[PSG]} Mikhail Aleksandrovich Kukushkin (;  born 26 December 1987 ) is a Kazakh professional tennis player of Russian descent . 
    & What is the nationality of the player in the 2010 President 's Cup who was born on 26 December 1987 ? 
    \\
    % \vspace{0.15cm}
    \texttt{[TAB]} \texttt{[TITLE]} Washington House of Representatives \texttt{[SECTITLE]} Composition -- Members ( 2019-2021 , 66th Legislature ) \texttt{[DATA]} District is 7. Position is 1. Representative is Jacquelin Maycumber. Party is Republican. Residence is Republic. Counties Represented is Ferry Okanogan Pend Oreille Spokane Stevens. First elected is 2017. \texttt{[PSG]} 
    & What is the residence of the Washington House of Representatives representative who was first elected in 2017. 
    \\
    \texttt{[TAB]} \texttt{[TITLE]} 1961 NFL expansion draft \texttt{[SECTITLE]} Player selections \texttt{[DATA]} Player is Don Joyce. Position is Defensive end. College Team is Tulane. Original NFL Team is Baltimore Colts. \texttt{[PSG]} Don Joyce ( October 8 , 1929 - February 26 , 2012 ) was an American football defensive end and professional wrestler.
    &  What was the original NFL team of the 1961 NFL expansion draft player who died on February 26 , 2012 ? 
    \\
    \texttt{[TAB]} \texttt{[TITLE]} 2009 Formula One World Championship \texttt{[SECTITLE]} Results and standings -- Grands Prix \texttt{[DATA]} Round is 2. Grand Prix is Malaysian Grand Prix. Pole position is Jenson Button. Fastest lap is Jenson Button. Winning driver is Jenson Button. Winning constructor is Brawn Mercedes. Report is Report. \texttt{[PSG]} The Malaysian Grand Prix was an annual auto race held in Malaysia . It was part of the Formula One World Championship from 1999 to 2017 and it was held during these years at the Sepang International Circuit . The first Malaysian Grand Prix was held in 1962 in what is now Singapore . 
    & What is the name of the constructor that won the 2009 Formula One World Championship round that was held at the Sepang International Circuit ? 
    \\
    \hline
    \end{tabular}
%   }
  \caption{Examples of synthesized corpus for pre-training. The queries are generated by a fine-tuned BART generotor given the input of flattened table segment. The generated questions not only are fluent and natural, but also consider mixed-modality information from tables and passages. }
  \label{tab:bart-example}%
\end{table*}

\section{Performance Analysis}

\subsection{Top-$k$ Retrieval Results} \label{app:topk-retrieval}
\begin{figure}[h]
     \centering
     \includegraphics[width=0.475\textwidth]{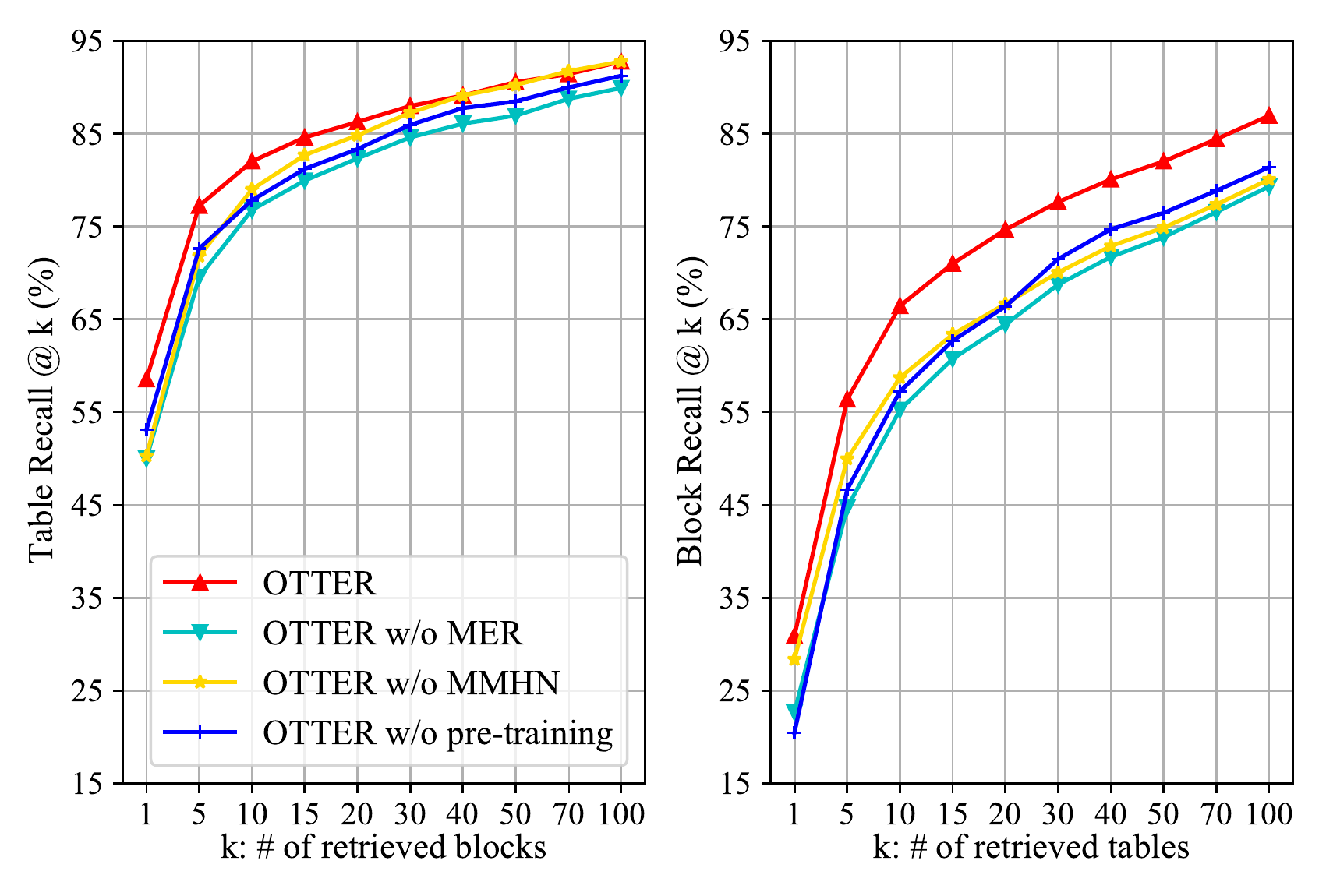}
     \caption{Top-k retrieval performance of retrievers on the dev set of OTT-QA. Full \textsc{OTTeR} substantially surpasses the other models in block recall, and in table recall when $k \leq 50$.}
     \label{fig:recall-result}
\end{figure}

Here, we show the detailed retrieval results of \textsc{OTTeR} with different components in Figure \ref{fig:recall-result}. 
The table recall at top-$k$ ranks and block recall at top-$k$ ranks are reported. 
We can find that full \textsc{OTTeR} substantially surpasses the models of other settings in block recall, and in table recall when $k \leq 50$. 

\subsection{Entity Linking} \label{app:entity-linking}
To understand the effects of entity linking, we evaluate the standalone entity linking accuracy and the retrieval performance.
We consider the following linking models: (1) GPT-2 used in \citet{chen2020ottqa}, which first augments the cell value by the context with a GPT-2 \cite{radford2019gpt2} and then uses BM25 to rank the blocks to the augmented form, (2) BLINK \cite{wu2019blink} used in \textsc{OTTeR}, which applys a bi-encoder ranker and cross-encoder re-ranker to link Wikipedia passages to the entities in flattened tables, (3) Oracle linker, which uses the original linking passages in the table. 

We evaluate the entity linking of on the OTTQA dev.
set following the settings in \citet{chen2020ottqa} and report the table-segment-wise F1 score. 
Table \ref{app:tab:entity-linking} shows the performance. 
We find that the F1 score of BLINK is higher than GPT-2, which leads to more relevant passages for tables. 

We further evaluate the retrieval performance with table-text corpus constructed by different entity linkers. 
Comparing GPT-2 and BLINK, we can find that the retrieval performance improves with the increased linking F1, especially when evaluated in block recall. 
The result indicates the importance of sufficient context information.

\begin{table}[t]
% \small
  \centering
  \resizebox{0.48\textwidth}{!}{
    \begin{tabular}{l|c|ccc|ccc}
    \hline
    & Linking & \multicolumn{3}{c|}{Table Recall} & \multicolumn{3}{c}{Block Recall} \\
    % \hline
    Linker & F1 & R@1  & R@10 & R@100 & R@1  & R@10 & R@100 \\
    \hline
    GPT2 & 50.4 & 58.2 & 81.5 & 92.5 &  28.6 & 64.0 & 83.7\\
    BLINK & 55.9 & 58.5  & 82.0  & 92.8  & 30.9  & 66.4  & 87.0 \\
    Oracle & 100 & 60.5 & 83.5 & 93.9 & 35.3 & 71.5 & 88.5 \\
    \hline
    \end{tabular}
    }
  \caption{Entity linking and retrieval results of different linkers. }
  \label{app:tab:entity-linking}
\end{table}%

\subsection{Embedding Dimension} \label{app:embed-dimen}
To maximumly eliminate the impact of embedding dimension in modality-enhanced representation (MER), we add a new
ablation by concatenating three \texttt{[CLS]} vectors as block representations, (i.e., $\mathbf{b}=[\mathbf{h}_{\texttt{[CLS]}}; \mathbf{h}_{\texttt{[CLS]}}; \mathbf{h}_{\texttt{[CLS]}} ]$), and training in the same way as MER=First
(i.e., $\mathbf{b}=[\mathbf{h}_{\texttt{[CLS]}}; \mathbf{h}_{\texttt{[TAB]}}; \mathbf{h}_{\texttt{[PSG]}} ] $). 
The results in Table \ref{app:tab:dimension} show that using specific representations of each modality still brings more sufficient information than \texttt{[CLS]} after maximumly eliminating the dimension bias.

\begin{table}[htbp]
% \small
  \centering
  \resizebox{0.48\textwidth}{!}{
    \begin{tabular}{l|ccc|ccc}
    \hline
    & \multicolumn{3}{c|}{Table Recall} & \multicolumn{3}{c}{Block Recall} \\
    % \hline
    Model & R@1  & R@10 & R@100 & R@1  & R@10 & R@100 \\
    \hline
    MER=First  & 58.5  & 82.0  & 92.8  & 30.9  & 66.4  & 87.0 \\
    CLS  & 57.5  & 80.4  & 92.5  & 29.6  & 64.0  & 86.5 \\
    \hline
    \end{tabular}
    }
  \caption{Ablation results on retrieval of MER dimension. }
  \label{app:tab:dimension}
\end{table}%

\end{document}